\pdfoutput=1

\documentclass[11pt]{article}

\usepackage[preprint]{acl}

\usepackage{times}
\usepackage{latexsym}

\usepackage[T1]{fontenc}

\usepackage[utf8]{inputenc}

\usepackage{microtype}

\usepackage{amsmath}
\usepackage{amsfonts}
\usepackage{nicefrac}

\usepackage{inconsolata}

\usepackage{enumitem}
\setlist{nolistsep}

\usepackage{graphicx}
\usepackage{subcaption}

\usepackage{cleveref}

\title{Synergy: End-to-end Concept Model}

\author{
    \textbf{Keli Zheng} \\
    Institute of Software, Chinese Academy of Science\\
    \texttt{zhengkl@ios.ac.cn}\\
    \and
    \textbf{Zerong Xie}\\
    The University of Hong Kong\\
    \texttt{zerong@connect.hku.hk}
}

\begin{document}

\maketitle

\begin{abstract}
    In this paper, we present \textit{Synergy}, a language model that bridges different levels of abstraction in an end-to-end fashion through a learned routing mechanism.
    Focusing on low-level linguistic abstraction, we trained our model as a byte-level language model.
    Our model spontaneously learns to tokenize bytes, producing fewer concept tokens than Byte-level Byte Pair Encoder (BBPE) tokenizers whiling keeping the comparable performance.
    By comparing with \textit{Llama3}, we observed an advantage of \textit{Synergy} under the same model scale and training dataset size.
    Further studies show that the middle part (the higher abstraction part) of our model performs better when positional encodings are removed, suggesting the emergence of position-independent concepts.
    These findings demonstrate the feasibility of tokenizer-free architectures, paving the way for more robust and flexible pipelines.
\end{abstract}

\section{Introduction}
\label{sec:introduction}

In the last few years, large language models (LLMs) have shown dazzling capabilities and enormous potential to solve a wide range of tasks. Many companies and research organizations have released their own large language models, such as ChatGPT, Llama, DeepSeek, etc., which have profoundly changed the way we work and live. Most of these products share a similar principle: a decoder-only-transformer-like model fed on a large number of tokens and trained with the predict-next-token task. Although researchers today have further developed sophisticated model structures and cleverer training approaches, the aforementioned principle still remains the core of most LLMs.

However, as~\citet{lcmteam2024largeconceptmodelslanguage} pointed out, such a core principle results in a drawback: it cannot efficiently process information at different levels of abstraction. Since the model is trained at the token level, it tends to ``think'' at the token level. However, many high-level abstract concepts are difficult to accurately describe in existing languages. It may prevent the model from performing better in scenarios where higher-level abstractions are critical, such as outlining a presentation, planning a tour, engineering a complicated program, etc. Many researchers believe that the ability to handle high-level abstract concepts can be spontaneously developed through training due to the multilayer structure of the transformer model. But a properly designed neural network structure can lead to a more efficient processing of high-level abstract concepts.

Creatively, in the Large Concept Model (LCM)~\citep{lcmteam2024largeconceptmodelslanguage} an autoencoder-based sentence embedding model is used to abstract token-level embeddings into sentence-level embeddings, and an encoder-only transformer model is fed with sentence-level embeddings and trained to predict next sentence embedding. Although their experiments have shown initial success, their solutions have an unavoidable disadvantage: The sentence embedding model is trained separately from the transformer model and have a quite different training target (in this case, as an autoencoder, to restore the original input), therefore the abstracted information may not be useful for the ultimate goal (in this case, to predict the next sentence embedding), leading to a low efficiency.

To overcome this disadvantage, we propose an end-to-end model \textit{Synergy}, where the encoding/decoding part and the processing part are trained together. The main challenge to train an end-to-end model is that the aforementioned two parts are working on token sequences with different lengths, which does not fit in a conventional decoder-only transformer. An encoder-decoder transformer is also not applicable due to the temporal dependency within/between these two sequences. Inspired by Mixture of Depths (MoD)~\citep{raposo2024mixtureofdepthsdynamicallyallocatingcompute}, we bridge those sequences using a router mechanism. Further, we configured those parts correspondingly to encourage a ``division of labor'' across different levels of abstraction, resulting in better efficiency.

Tokens are also concepts at a low level of abstraction, except that they have already been symbolized in our language. Using statistic-based tokenizers like Byte-level Byte Pair Encoding (BBPE) tokenizers, we already achieved a quite efficient abstraction from characters/bytes to tokens/words. However, with the attempts to process multimodal data such as images, audio, video, etc. with LLMs, the ability to process data in low-level of abstractions becomes important.

Considering the fact that low-level abstraction is relatively easier, and therefore consumes affordable resources, our experiments focused on low-level abstraction tasks. We compared our model \textit{Synergy} with \textit{Llama3}, and observed an advantage under the same model scale and training dataset size. Further studies show that \textit{Synergy} can generate fewer tokens than BBPE tokenizers. In addition, we observed an unexpected decrease of loss when removing positional encoding from the middle part, suggesting the emergence of position-independent concepts.

\textbf{In short, our contributions are:}
\begin{itemize}
    \item We proposed an end-to-end model \textit{Synergy} that bridges different levels of abstraction.
    \item We conducted experiments to show the efficiency and feasibility of tokenizer-free \textit{Synergy} as a byte-level language model.
    \item We uncovered the emergence of position-independent concepts in the middle part of \textit{Synergy} by removing positional encoding.
\end{itemize}

\section{Methodology}
\label{sec:methodology}

The main idea of \textit{Synergy} is to reduce the number of tokens using a routing mechanism, then process the resulting shorter sequence using a decoder-only transformer. Like a normal decoder-only transformer, the whole model consumes a sequence of tokens as input, produces a corresponding sequence of tokens as output, and is trained by predict-next-token tasks.

\begin{figure*}[h]
    \centering
    \begin{subfigure}[b]{0.2\textwidth}
        \centering
        \includegraphics[height=5.5cm]{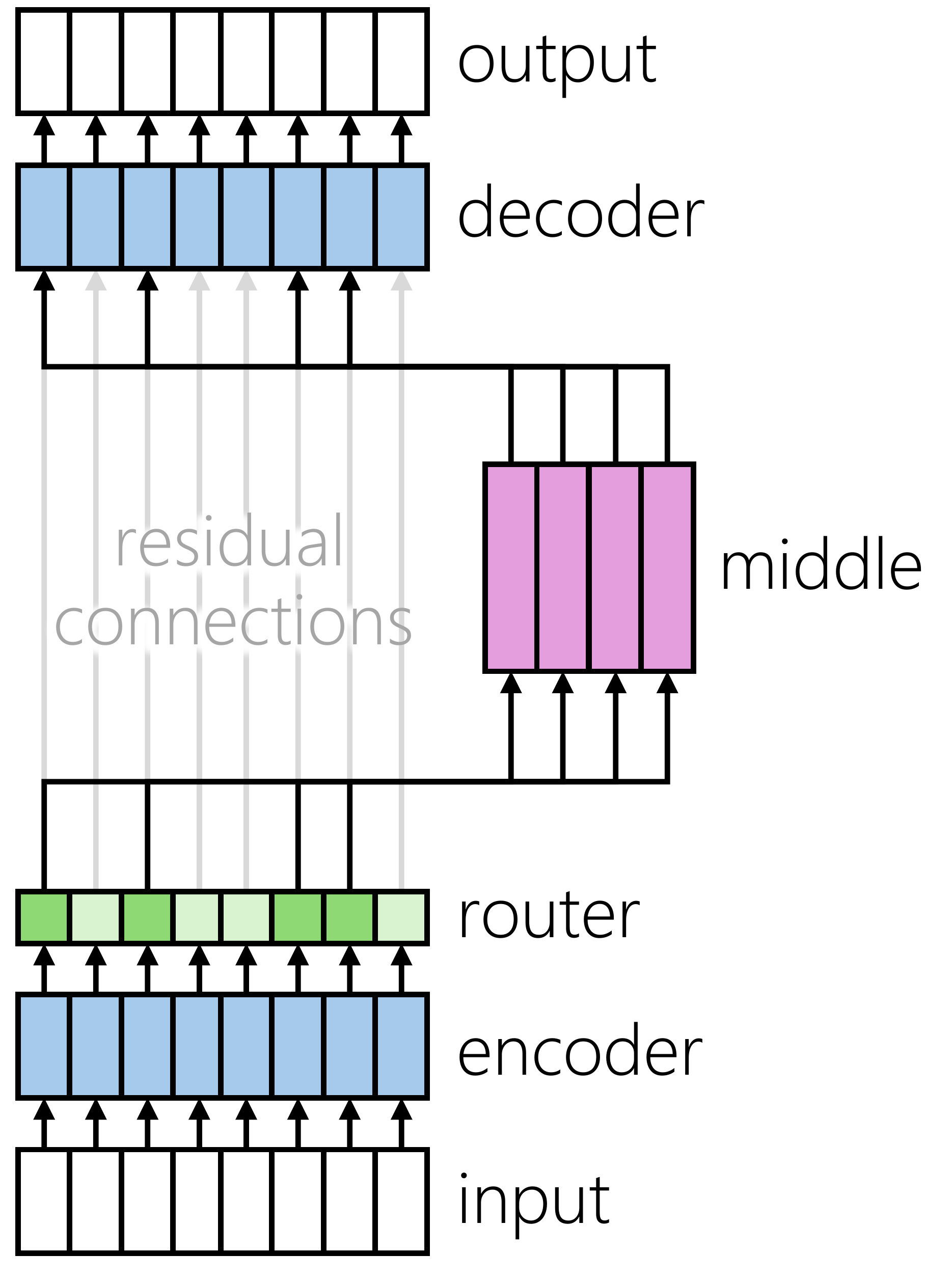}
        \caption{The overview}
        \label{fig:overview}
    \end{subfigure}
    \hspace{1.5cm}
    \begin{subfigure}[b]{0.2\textwidth}
        \centering
        \includegraphics[height=6cm]{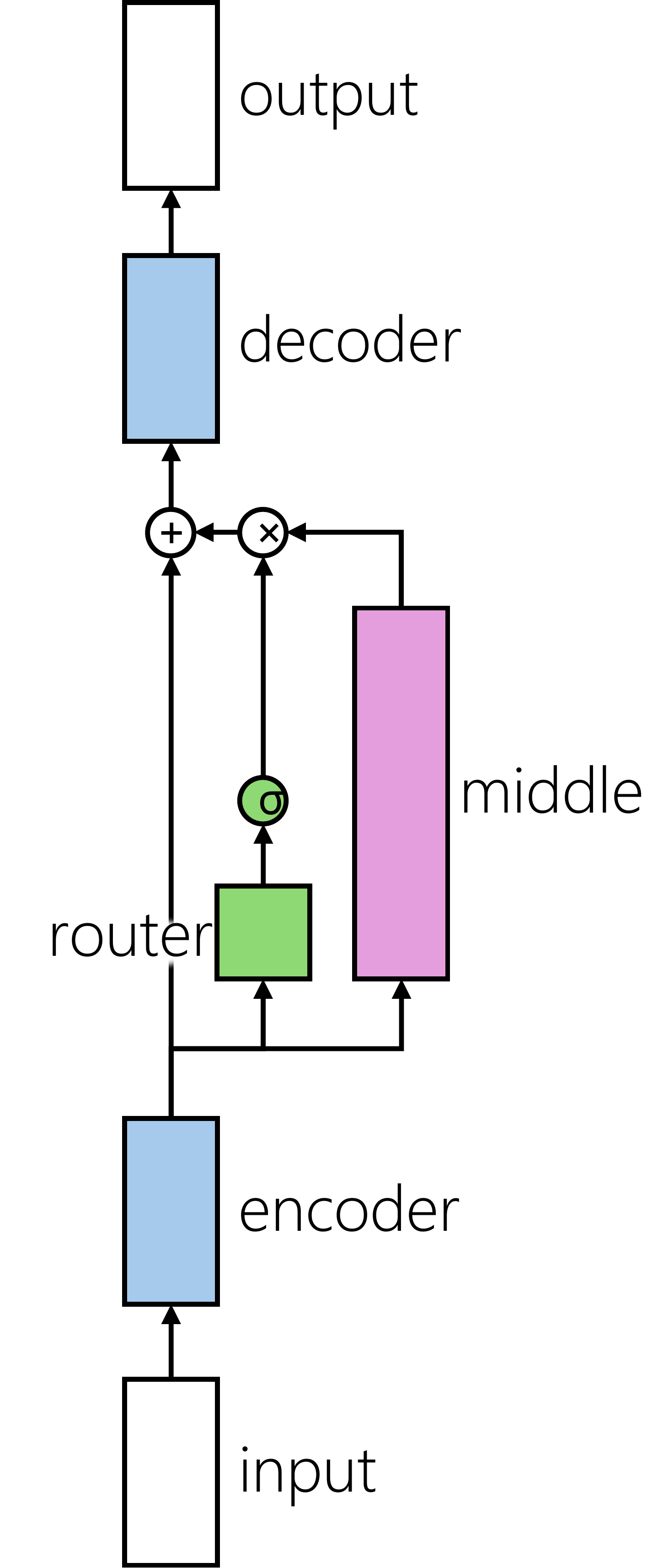}
        \caption{Routing (active)}
        \label{fig:framework_active}
    \end{subfigure}
    \hspace{0.5cm}
    \begin{subfigure}[b]{0.2\textwidth}
        \centering
        \includegraphics[height=6cm]{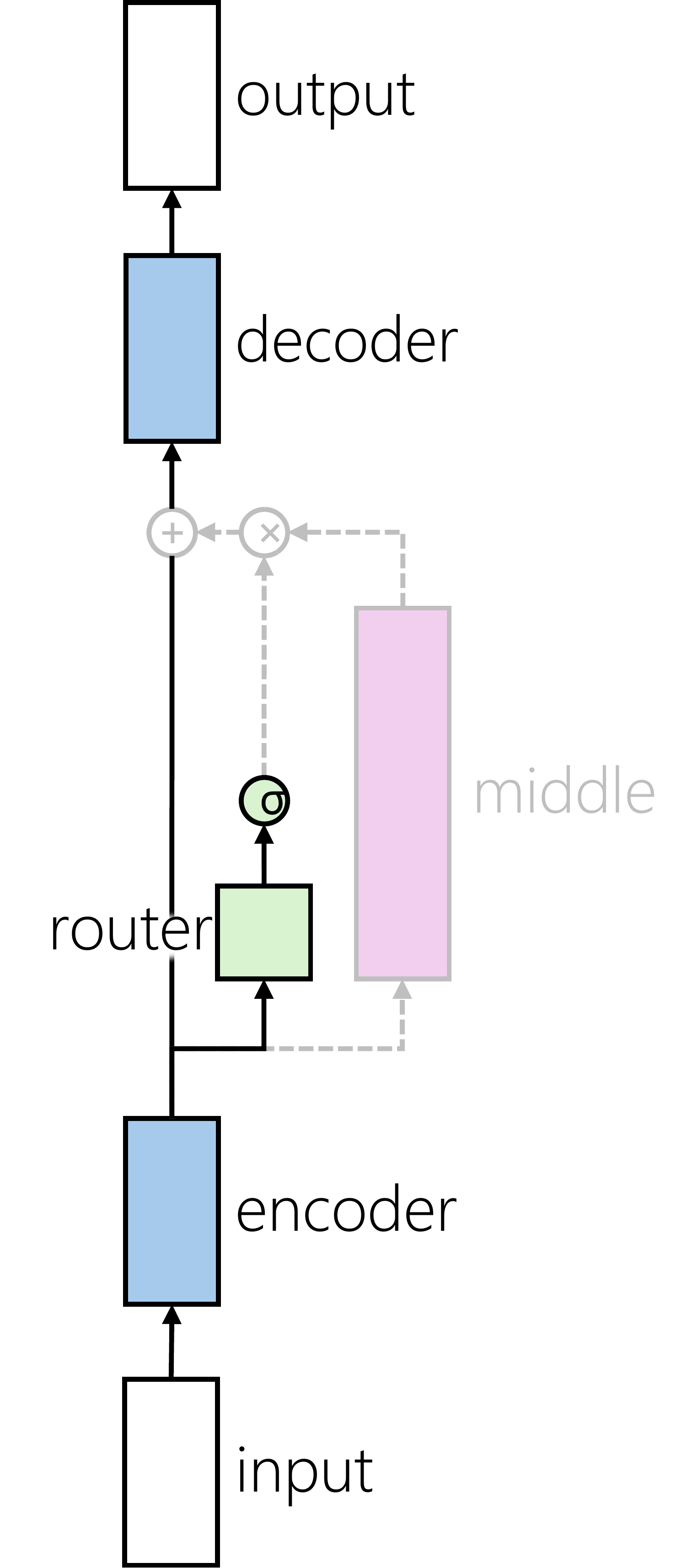}
        \caption{Routing (inactive)}
        \label{fig:framework_inactive}
    \end{subfigure}
    \caption{The architecture of Synergy}
    \label{fig:architechture}
\end{figure*}

As it shows in \cref{fig:overview}, the model is split into three parts — encoder, middle, and decoder, which are all decoder-only transformers. The output from the encoder is passed to the decoder. For a portion of tokens, the encoder output also passes through the middle part and then adds onto the decoder input.

A router is responsible to determine which tokens can go through the middle part. As it shows in \cref{fig:framework_active,fig:framework_inactive}, the routing mechanism is similar to that in MoD~\citep{raposo2024mixtureofdepthsdynamicallyallocatingcompute}. A weight factor $w_i$ is computed from the encoder output $x_i$ through a simple linear layer for each token (indexed by $i$). A portion of tokens are picked to pass through the middle part via a top-k operation on the weight factor. The picking result is represented as a mask $m_i$. For the picked tokens, the middle output $\text{Middle}(x_i)$ is multiplied by a gating factor $\sigma_i$ before adding to the decoder input. The routing mechanism can be formally described as the following formulas:

\begin{align}
    w_i &= \text{Router}(x_i) \\
    m_i &= \text{topkmask}(w_i, k) \\
    \sigma_i &= \text{sigmoid}(w_i) \\
    y_i &= x_i + m_i\ \sigma_i\ \text{Middle}(x_i)
\end{align}

Where,
\begin{itemize}
    \item $x_i$ — The encoder output of the $i$-th token.
    \item $w_i$ — The weight factor of the $i$-th token computed by the router.
    \item $k$ — A hyperparameter that controls the number of tokens that can pass through the middle part.
    \item $m_i$ — The mask that represents whether the $i$-th token can pass through the middle part.
    \item $\sigma_i$ — The gating factor of the $i$-th token computed from $w_i$.
    \item $y_i$ — The decoder input of the $i$-th token.
\end{itemize}

Here we can consider the weight factor $w_i$ as an importance indicator of each token. While a token is important, it is reasonable to allocate more computation to process it, so it activates the middle part. As a result, the middle part processes fewer tokens than the encoder/decoder part, achieving our goal of token number compressing.

We assume that the encoder/decoder and middle parts work in different levels of granularity, for example, the encoder/decoder part should be responsible for concrete tasks, focusing on few tokens, while the middle part should take on abstract tasks, involving long-range context. So we used different configurations for them respectively. On the one hand, we use local attention in the encoder/decoder part. On the other hand, we removed the positional encoding in the middle part.

\section{Experiments}
\label{sec:experiments}

\subsection{Experimental Setup}
\label{subsec:experimental-setup}

\subsubsection{Dataset Configurations}
\label{subsubsec:dataset-configurations}

In all experiments, we use the English Wikipedia text corpus~\footnote{To be exact, we use the subset 20231101.en from \url{https://huggingface.co/datasets/wikimedia/wikipedia/viewer/20231101.en}}, where 70\% randomly sampled rows are used for training, remaining the rest for evaluation and testing. We chose this dataset because we found it best highlights the characteristics of our model.

To train our model \textit{Synergy} as a byte-level language model, we tokenize the dataset using a simple UTF8 tokenizer, which encodes any text into bytes according to the UTF-8 standard and then treats each byte as a token.

\subsubsection{Model Hyperparameters}
\label{subsubsec:model-hyperparameters}

In all experiments, our model \textit{Synergy} has totally 32 layers, with 4 layers in the encoder part, 24 layers in the middle, and 4 layers in the decoder part. The embedding dimension and the hidden dimension of all transformer layers, is 1024. Every transformer layer consists of a 16-head self-attention layer with 64-dimensional Q, K, V, and an MLP layer with SwiGLU activation~\citep{shazeer2020gluvariantsimprovetransformer} and 4096-dimension intermediate~\footnote{The $\frac{2}{3}$ factor introduced by SwiGLU is \textbf{not} multiplied.}.

By default, the model is trained with sequences of length 1024, and the corresponding number of tokens in the middle part is 224. For positional encoding, we choose the widely used RoPE~\citep{su2023roformerenhancedtransformerrotary}. Recall that we deliberately removed the positional encoding in the middle part, so it only affects the encoder and decoder.

\subsubsection{Hardware and Costs}
\label{subsubsec:hardware-and-costs}

We finished all experiments on a single machine with one GPU (NVIDIA H20 96GB). The whole training set consists of about $12$G bytes. Using our experimental configurations, the whole training phase of \textit{Synergy} lasts for about $62$ hours (while it lasts for about $40$ hours for \textit{Llama3}).

\subsection{Comparison between \textit{Synergy} and \textit{Llama3}}
\label{subsec:comparison}

To evaluate the ability of our model to abstract word-level concepts, we arranged a comparison between \textit{Synergy} and \textit{Llama3}. For \textit{Llama3} we use the official tokenizer with $128$k words/phrases in its vocabulary, while \textit{Synergy} is equipped with a simple UTF8 tokenizer as mentioned above.

To ensure a fair comparison, a number of special configurations are used:

\begin{itemize}
    \item Since two models are using different tokenizers, the perplexity is no longer a suitable metric for this comparison. Follow previous works~\citep{10.1162/tacl_a_00461,yu2023megabytepredictingmillionbytesequences,wang2024mambabytetokenfreeselectivestate,pagnoni2024bytelatenttransformerpatches}, we use Bits-Per-Byte (BPB), which is independent of tokenizer, as the metric for comparison.
    \item Since two models are using different tokenizers, it becomes tricky to ensure an equal context size for both models. To avoid this unfairness, we configured a long enough context size for both models correspondingly (\textit{Synergy}: $1024$ bytes; \textit{Llama3}: $224$ tokens). Then we clipped the training text samples into shorter independent segments so that one segment can be fit into the context after tokenization for both models. It is fair for both models because they are trained with the same text segments without clipping.
    \item To ensure the fairness in model scale, we use 32 layers in the \textit{Llama3}, which is consistent with the total number of layers in \textit{Synergy}, and use the same set of hyperparameters for all transformer layers. However, there is still a difference in the number of parameters (\textit{Llama3}: $0.8$B, \textit{Synergy}: $0.5$B), which is caused by the embedding layer: while \textit{Synergy} has a tiny vocabulary ($256$ plus special tokens), \textit{Llama3} has a large vocabulary (upto $128$k), requiring extra $0.3$B parameters for token embedding. Despite this difference, we still consider two models matched, as they have the same number of transformer layers.
\end{itemize}

\begin{figure}[h]
    \centering
    \includegraphics[width=7.5cm]{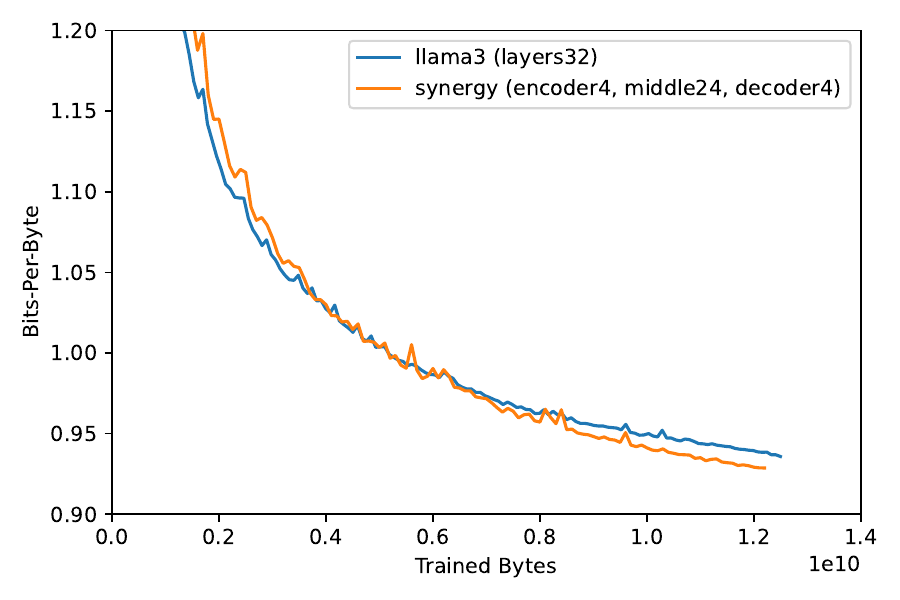}
    \caption{BPB curves of both models}
    \label{fig:bpb-curves}
\end{figure}

As shown in~\cref{fig:bpb-curves}, our model yields a better result when the training data exceeds $0.6$T bytes, which indicates an advantage of our model compared to \textit{Llama3}. However, the price of getting this advantage is about 1.5 times more computation time, see~\cref{subsec:extra-flops} for more details.

\subsection{Positioning Modes}
\label{subsec:positioning-modes}

One of the special designs of our model is the removal of positional encoding for the middle part. It originates from a series of experiments with different positioning modes for the middle part. We can roughly divide them into three categories:

\begin{itemize}
    \item Original: use the original position picked from the input.
    \item Sigma: use the accumulated value of the gating factor $\text{cumsum}(\sigma_i)$ as the position.
    \item None: do not use any positional encoding.
\end{itemize}

For ``Sigma'', there are variants. We can accumulate gating factors of all tokens or only those of picked tokens. And we can remain or cut the gradient trace of $\sigma_i$ when computing the positional encodings. In total, we have 4 variants of ``Sigma''.

\begin{figure}[h]
    \centering
    \includegraphics[width=7.5cm]{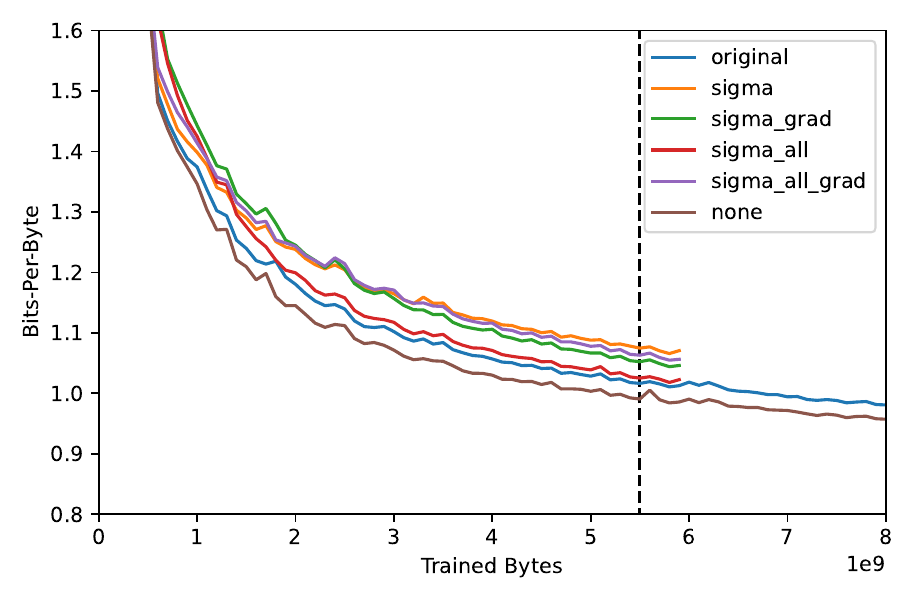}
    \caption{BPB curves of different positioning modes}
    \label{fig:positioning-modes}
\end{figure}

We trained our model correspondingly with all 6 modes. Considering the training cost, we stopped most training experiments at $6$T tokens, where the trend is stable enough. As we can see in \cref{fig:positioning-modes}, the ``None'' mode is the best among them. We collected BPB values and listed them in \cref{tab:positioning-modes}.

\begin{table}
    \centering
    \begin{tabular}{lc}
        \hline
        Positioning Mode & BPB (at $5.5$T token) \\
        \hline
        original         & 1.0164                \\
        sigma            & 1.0747                \\
        sigma\_grad      & 1.0523                \\
        sigma\_all       & 1.0251                \\
        sigma\_all\_grad & 1.0630                \\
        \textbf{none}    & \textbf{0.9906}       \\
        \hline
    \end{tabular}
    \caption{BPB values of different positioning modes}
    \label{tab:positioning-modes}
\end{table}

This phenomenon is quite above our expectations. We assume that this is because the concept in the encoder output is actually independent of concrete positions, and thus, in contrast, it is interfered with by positional encodings. It does not mean that the concept is completely independent of position. The position information about can be extracted in the encoder part and absorbed into the concept. Such a phenomenon also reveals a potential — as the middle part does not rely on the positional encoding, it may have better extrapolation performance.

\subsection{Concept Tokens Number}
\label{subsec:middle-tokens-number}

\begin{figure*}[h]
    \centering
    \includegraphics[width=14cm]{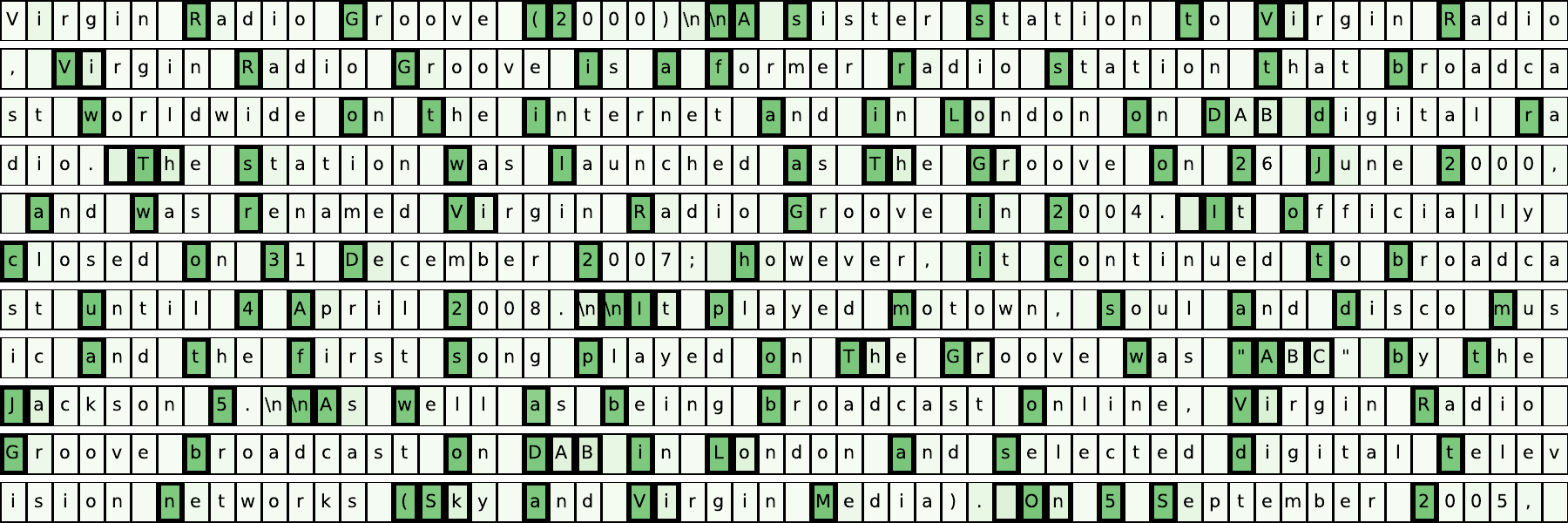}
    \caption{Visualization of router output, where green color represents the weight factor (the darker green, the greater weight) and the black border represents the routing mask (bold border means passing through the middle part).}
    \label{fig:routing-visualization}
\end{figure*}

To inspect how Synergy tokenizes bytes, we visualized the output of the router. As we can see in \cref{fig:routing-visualization}, the weight factor shows an obvious relation with word separations, which means that our model somehow spontaneously learns the ability to segment the bytes into words.

In addition, we observed that a number of bytes are picked with low weight factors. This indicates a redundancy in tokenization. These bytes have low weight factors and correspondingly have low gating factors, so that we can actually drop these tokens from the middle part without significant impact on the final output.

\begin{figure}[h]
    \centering
    \includegraphics[width=7.5cm]{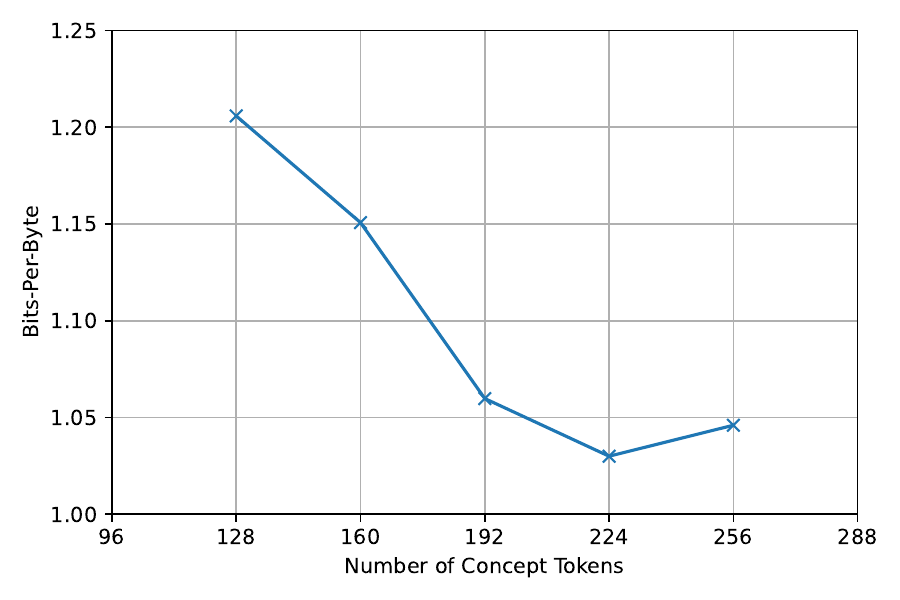}
    \caption{The influence of concept tokens counts to BPB (collected at $4$T tokens).}
    \label{fig:middle-tokens}
\end{figure}

To further investigate this redundancy, we conduct a series of experiments to test the influence of the concept token count (i.e., the number of tokens processed in the middle part) on the performance. As shown in \cref{fig:middle-tokens}, the performance does not decrease until the number of concept tokens falls below 192, while the number of tokens from \textit{Llama3} tokenizer is about $1024/4.25=240.94$. This result shows the potential of our model to tokenize bytes more efficiently, yielding fewer concept tokens than BBPE tokenizers.

\section{Related Works}
\label{sec:related-works}

Prior to our work, there have been a number of works related to the abstraction of information, some of which has inspired our design. We organized the related works in the following subsections classified by their approaches.

\subsection{Sparse Attention}
\label{subsec:sparse-attention}

Sparse attention is a powerful technique to reduce the computational cost of attention, especially in long-context scenarios.

In the early stages, there exists a group of sparse attention methods which rely on predefined structures like fixed strides~\citep{child2019generatinglongsequencessparse}, dilated sliding window~\citep{beltagy2020longformerlongdocumenttransformer} and attention sinks~\citep{xiao2024efficientstreaminglanguagemodels}. The sparsity in these methods has little correlation with the content, which may limit their efficiency.

To address this issue, a group of dynamic sparse attention mechanisms were proposed. The representative ones are H2O~\citep{zhang2023h2oheavyhitteroracleefficient}, Quest~\citep{tang2024questqueryawaresparsityefficient}, Minference~\citep{jiang2024minference10acceleratingprefilling} and RetrievalAttention~\citep{liu2024retrievalattentionacceleratinglongcontextllm}. These methods significantly reduced the computational cost by maintaining a subset of KV pairs in long-context inference. But these methods cannot be applied to the training phase.

In recent years, new approaches, such as ~NSA~\citep{yuan2025nativesparseattentionhardwarealigned} and MoBA~\citep{lu2025mobamixtureblockattention}, introduce content-dependent block-wise sparsity to reduce the computational cost of attention. These methods can speed up both training and inference phases. However, they still cannot easily handle abstract information that diffuses throughout the long sequence (e.g., ambient gradients, style transformation).

Moreover, the improvement of attention does not change the fact that the whole model still operates at the token level, and is challenging to process abstract concepts that cannot be clearly described in words.

\subsection{RNN-like language model}
\label{subsec:rnn-like-language-model}

Except for the aforementioned efficient attention, there are other approaches that break the fundamental principle of attention, resulting in a number of RNN-like language models. Starting with Linformer~\citep{wang2020linformerselfattentionlinearcomplexity}, a number of attempts like RWKV~\citep{peng2023rwkvreinventingrnnstransformer,peng2024eaglefinchrwkvmatrixvalued}, Mamba~\citep{gu2024mambalineartimesequencemodeling,dao2024transformersssmsgeneralizedmodels}, and RetNet~\citep{sun2023retentivenetworksuccessortransformer} have made significant progress.

However, despite the high efficiency of these models in dealing with long-context tasks, they remain confined to a single level of abstraction, in most cases, the level of words. We note that in some of these rnn-like models, state transition with different frequencies is deliberately involved to handle information of different granularity, which corresponds to our idea. But instead of processing everything in a sequence at the finest granularity, it would be more efficient to separately process sequences of granularity.

\subsection{Hierarchical transformers}
\label{subsec:tokenization-&-embedding}

Hierarchical transformers, as a promising approach to process long sequences, have been studied in many previous works. Hourglass~\citep{nawrot2022hierarchicaltransformersefficientlanguage}, MegaByte~\citep{yu2023megabytepredictingmillionbytesequences} and Block Transformer ~\citep{ho2024blocktransformerglobaltolocallanguage} are proposed with a hierarchical transformer architecture with fixed-size pooling and upsampling mechanisms. However, fixed block size limits its adaptability to various situations.

MrT5~\citep{kallini2025mrt5dynamictokenmerging} introduces an efficient variant of ByT5 that integrates a dynamic token deletion mechanism, but it is not applicable to decoder-only transformers. In the meantime, Dynamic-Pooling Transformer~\citep{nawrot-etal-2023-efficient,ahia2024magnetimprovingmultilingualfairness} introduced a boundary predictor trained by Gumbel-sigmoid stochastic re-parameterization. Their idea corresponds to ours, but uses a different approach to train the router. And another difference is that our model has no deliberate pooling/upsampling because we do not presuppose the task to be word segmenting, but token filtering/compressing.

Moreover, BLT~\citep{pagnoni2024bytelatenttransformerpatches} employed an entropy-based segmenting mechanism to segment the long sequence into patches (this approach is also mentioned in Dynamic-Pooling Transformer~\citep{nawrot-etal-2023-efficient}). LCM~\citep{lcmteam2024largeconceptmodelslanguage} uses a sentence embedding model to extract sentence-level embeddings from long sequences and then process them with a predict-next-sentence-embedding decoder-only transformer, making good progress in abstracting sentence-level concepts. These methods have initially inspired our designs.

\section{Limitations}
\label{sec:limitations}

\subsection{Instability}
\label{subsec:instability}

We observed the instability during the training of our model. For example, at the beginning phase of the training, we observe glitches (a sudden spike in BPB) from time to time. In some cases, the accuracy converges to a higher value after the glitch, resulting in a poor training result. Even if there is no glitch, each training session may also yield very different BPB values, even using the same hyperparameters.

We hypothesize that this is caused by the imperfect router training method. Since there is no direct gradient connection from token sampling probabilities to the loss (recall that the sampling is operated by topK), the change of picking rule is unstable. On the one hand, breaking changes can be committed during training, resulting in glitches. On the other hand, beneficial changes may fail to occur, causing a poor result. One of our future works is to explore better ways to train the router both stably and cost efficiently.

Considering this limitation, the experimental results presented in this paper are partial results after being filtered out the poor outliers. Due to the high costs of pretraining a model from scratch, yet we were unable to repeat all experiments multiple times. But all results are confirmed to be reproducible.

\subsection{Extra FLOPs}
\label{subsec:extra-flops}

In our experiment, \textit{Synergy} consumes about $1.5$ times more FLOPs than \textit{Llama3}. But it is worth noting that the extra FLOPs do not come from attention over long sequences, but from the MLP layers in encoder/decoder that inevitably run on each byte. Considering that every token has about $4.25$ bytes on average under our experimental setting, such extra FLOPs are reasonable. In long-context scenarios, where the attention dominates the FLOPs consumption, this disadvantage will be mitigated. On the other hand, considering that the encoder and decoder are using local attention, a proper cache-aware implementation may greatly improve its performance.

\section{Conclusion}
\label{sec:conclusion}

In this paper, we proposed an end-to-end model \textit{Synergy} that bridges different levels of abstraction through a learned routing mechanism. We trained \textit{Synergy} as a byte-level language model and compared it with \textit{Llama3}, which uses a BBPE tokenizer, and observed an advantage of our model over \textit{Llama3} under the same model scale and training dataset size. Further studies show that our model can compress bytes into fewer concept tokens than BBPE tokenizers, which reflects its efficiency in abstracting word-level concepts. Moreover, experiments on different positioning options for the middle part yield an unusual result: the model works best when the positional encoding is removed. This phenomenon shows that the concepts processed in the middle part are somehow position-independent. Our findings demonstrate the feasibility of tokenizer-free architectures, paving the way for more robust and flexible pipelines.

However, our model is far from perfect. It suffers from instability in training and extra computational cost. It leaves much to explore for the future, such as exploring better ways to train the router, fine-tuning the layers of each part for better efficiency and using specialized structures for encoder, middle, and decoder parts. In addition, it is worth investigating the extrapolation performance of the middle part in the absence of positional encoding. It is also worth a try to use our model to abstract sentence-level concepts.

\bibliography{references}

\end{document}